\documentclass[11pt,a4paper]{article}
\usepackage[hyperref]{ranlp2021}
\usepackage{times}
\usepackage{latexsym}

\usepackage{microtype}

\usepackage{times}
\usepackage{latexsym}
\usepackage{amsmath}
\usepackage{hyperref}
\usepackage{multirow}
\usepackage{makecell}

\usepackage{microtype}
\usepackage{graphicx}
\usepackage{enumerate}

\aclfinalcopy 

\title{Cross-lingual Offensive Language Identification for Low Resource Languages: The Case of Marathi}

\author{Saurabh Gaikwad\textsuperscript{1}, Tharindu Ranasinghe\textsuperscript{2}, Marcos Zampieri\textsuperscript{1}, Christopher M. Homan\textsuperscript{1}  \\
  \textsuperscript{1}Rochester Institute of Technology, USA \\
  \textsuperscript{2}University of Wolverhampton, UK \\
  \texttt{T.D.RanasingheHettiarachchige@wlv.ac.uk} \\}

\date{}

\begin{document}
\maketitle
\begin{abstract}
The widespread presence of offensive language on social media motivated the development of systems capable of recognizing such content automatically. Apart from a few notable exceptions, most research on automatic offensive language identification has dealt with English. To address this shortcoming, we introduce {\em MOLD}\footnote{MOLD is available at: \url{https://github.com/tharindudr/MOLD}}, the Marathi Offensive Language Dataset. {\em MOLD} is the first dataset of its kind compiled for Marathi, thus opening a new domain for research in low-resource Indo-Aryan languages. We present results from several machine learning experiments on this dataset, including zero-short and other transfer learning experiments on state-of-the-art cross-lingual transformers from existing data in Bengali, English, and Hindi.
\end{abstract}

\section{Introduction}

The presence of hate speech, cyber-bullying, and other forms of offensive language in online communities is a global phenomenon.  
Even though thousands of languages and dialects are widely used in social media, most studies on the automatic identification of such content consider English only, a language for which datasets and other resources such as pre-trained models exist \cite{rosenthal2020}. In the past few years researchers have studied this problem on languages such as Arabic \cite{mubarak2020arabic}, French \cite{chiril-etal-2019-multilingual}, and Turkish \cite{coltekin2020} to name a few. In doing so, they have created new datasets for each of these languages. Competitions such as OffensEval \cite{zampieri-etal-2020-semeval} and TRAC \cite{kumar-etal-2020-evaluating}
provided multilingual datasets, which enabled the use of data augmentation methods \cite{ghadery-moens-2020-liir}, multilingual word embeddings \cite{pamungkas2019cross}, and cross-lingual contextual word embeddings \cite{ranasinghe-etal-2020-multilingual} to tackle this problem.

In this paper, we revisit the task of offensive language identification for low resource languages, focusing on Marathi, an Indo-Aryan language spoken by over 80 million people, most of whom live in India. Even though Marathi is spoken by a large population, it is relatively low-resourced compared to other languages spoken in the region. We collect and annotate the first Marathi offensive language identification dataset to date and we train a number of monolingual models on this dataset. Finally, we explore state-of-the-art cross-lingual learning methods to project predictions to Marathi from Bengali, Hindi, and English. 
We address two research questions in this paper:\\

\noindent \textbf{RQ1}: What is the impact of the dataset size in monolingual and cross-lingual models for offensive language identification? While the Marathi dataset is relatively small, cross-lingual transfer learning methods allow us to take advantage of larger available datasets in other languages. \\

\noindent \textbf{RQ2}: What is the influence of language similarity in cross-lingual predictions for offensive language identification? Previous work used English as the base language to make predictions in lower resourced languages. 
In this paper we use two Indo-Aryan languages, Bengali and Hindi, to project predictions into Marathi. \\

\noindent Our main contributions are the following:

 \begin{enumerate}
 \item We release {\em MOLD}, the Marathi Offensive Language Dataset, with nearly 2,500 annotated tweets. {\em MOLD} is the first dataset for offensive language identification in Marathi.

 \item We evaluate the performance of several traditional machine learning models (e.g. SVMs) and deep learning models (e.g. LSTM) trained on {\em MOLD}.

 \item We apply cross-lingual transformers to offensive language identification in Marathi. We take advantage of existing data in English and in two Indo-Aryan languages, Hindi and Bengali, to project predictions to Marathi and we compare the results of these strategies. 
 To the best of our knowledge, this is the first paper to study closely-related languages in transfer learning for offensive language identification. 

\item In addition to {\em MOLD}, we make the code and the models freely available to the community.

\end{enumerate}

\section{Related Work}
\label{sec:RW}

The problem of offensive content online has been widely studied using computational models. Researchers have trained system to recognize various types of such content such as {\em cyberbulling}, {\em hate speech}, and many others. In terms of computational approaches, early studies approached the problem using feature engineering and classical machine learning classifiers, most notably SVMs \cite{dadvar2013improving,malmasi2017detecting}, while more recent work applied deep neural networks combined with word embeddings \cite{aroyehun2018aggression, hettiarachchi-ranasinghe-2019-emoji}. With the development of large pre-trained transformer models such as BERT and XLNET \cite{devlin2019bert,NEURIPS2019_dc6a7e65}, several studies have explored the use of general pre-trained transformers in offensive language identification \cite{liu-etal-2019-nuli,ranasinghe2019brums,bucur-etal-2021-exploratory} as well retrained or fine-tuned models on offensive language corpora such as HateBERT \cite{caselli2020hatebert}.

While the vast majority of studies address offensive language identification using English data \cite{yao2019cyberbullying,ridenhour2020detecting}, several recent studies have created new datasets for various languages and applied computational models to identify such content in Arabic \cite{mubarak2020arabic}, Dutch \cite{tulkens2016dictionary}, French \cite{chiril-etal-2019-multilingual}, German \cite{wiegand2018overview}, Greek \cite{pitenis2020}, Hindi \cite{bohra2018dataset}, Italian \cite{poletto2017hate}, Portuguese \cite{fortuna2019hierarchically}, Slovene \cite{fiser2017}, Spanish \cite{plaza2021comparing}, and Turkish \cite{coltekin2020}. A recent trend is the use of pre-trained multilingual models such as XLM-R \cite{conneau2019unsupervised} to leverage available English resources to make predictions in languages with less resources \cite{plaza2021comparing, ranasinghe-etal-2020-multilingual, ranasingheTALLIP, ranasinghemudes, sai-sharma-2021-towards}. 
This is made possible by the availability of the aforementioned datasets as well multilingual datasets made available at shared tasks such as HASOC 2019 \cite{mandl2019overview}, TRAC 2018 and 2020 \cite{kumar2018benchmarking,kumar-etal-2020-evaluating}, and two tasks at SemEval: HatEval 2018 \cite{basile2019semeval} and OffensEval 2020 \cite{zampieri-etal-2020-semeval}.




\section{Datasets}
\label{sec:data}

We present {\em MOLD} and four other datasets used in this work: the Bengali dataset \cite{bhattacharya-etal-2020-developing} used in the TRAC-2 shared task \cite{kumar-etal-2020-evaluating}---henceforth {\em BE}, the Hindi dataset \cite{mandl2019overview} used in the HASOC 2019 shared task---henceforth {\em HI}, and the English datasets used in OffensEval, SemEval-2019 Task 6 and SemEval-2020 Task 12---henceforth {\em EN-OLID}  \cite{OLID} and {\em EN-SOLID} \cite{rosenthal2020}, respectively. 

To annotate {\em MOLD}, we followed OLID's annotation scheme for English which has been replicated in SOLID and in datasets in Greek \cite{pitenis2020}, Turkish \cite{coltekin2020} and many other languages. OLID's taxonomy comprises the following three levels:\\

\vspace{-4mm}

\noindent {\bf Level A:} Offensive language identification: offensive (OFF) vs. non-offensive (NOT) \\

\vspace{-4mm}

\noindent {\bf Level B:} Categorization of offensive language: targeted insult or thread vs. untargeted profanity.\\

\vspace{-4mm}

\noindent {\bf Level C:} Offensive language target identification: individual vs. group vs. other.\\

\vspace{-4mm}

\noindent This hierarchical taxonomy represents multiple types of offensive content in a single annotation scheme (e.g. targeted insults to an individual are often {\em cyberbullying} and targeted insults to a group are often {\em hate speech}) making it a great fit for cross-lingual learning applied to low-resource languages like Marathi. We used OLID level A labels to annotate {\em MOLD} and we map these labels to those included in the Bengali and Hindi datasets. 

\paragraph{{\em MOLD}} The Marathi dataset contains data collected from Twitter using the Twitter API.
We aimed to achieve a similar distribution of offensive vs. non-offensive content present in OLID, which contains around 33\% offensive and 67\% non-offensive tweets. To make sure that both classes were represented, we used both offensive and non-offensive keywords. For the offensive content we used 22 common curse words in Marathi and for the non-offensive content we used search phrases related to politics, entertainment, and sports along with the hashtag \#Marathi. 

We collected a total 2,547 tweets that were annotated by 6 volunteer annotators who are native speakers of Marathi with age between 20 and 25 years old and a bachelors degree. The annotation task is a binary classification, in which annotators assigned tweets as offensive (OFF) or not offensive (NOT). The annotators could flag a tweet as invalid if it contained four or more non-Marathi words. The final version of {\em MOLD} contains 2,499 annotated tweets randomly split 75\%/25\% into training and testing sets, respectively. We used Cohen's kappa \cite{carletta1996assessing} to measure agreement between pairs of annotators. We provided a common set of 100 instances to each of the three pairs of annotators and we report scores of 0.91 between A1 and A2, 0.79 between A3 and A4, and 0.77 between A5 and A6. Table \ref{tab:number_instances} shows  dataset statistics, including class distribution.

\begin{table}[!ht]
\centering
\setlength{\tabcolsep}{4.5pt}
\scalebox{0.95}{
\begin{tabular}{lccc}
\hline
\bf Class   & \bf Training & \bf Testing & \bf Total \\ \hline

Not Offensive &1,205&418 & 1,623   \\
Offensive &669&207 &  876  \\ \hline
\bf Total &1,874&625  & 2,499    \\
 \hline

\end{tabular}
}
\caption{Number of instances and class distribution of NOT and OFF tweets in {\em MOLD}.}
\label{tab:number_instances}
\end{table}

\begin{table*}[!ht]
\centering
\setlength{\tabcolsep}{4.4pt}
\scalebox{.90}{
\begin{tabular}{llcccp{8.5cm}}
\hline
\bf Code & \bf Language & \bf Dataset  & \bf Instances & \bf Source & \bf Labels  \\ \hline

{\em BE} & Bengali & TRAC & 4,000 & F & overtly aggressive, covertly aggressive, non aggressive     \\
{\em EN-OLID} & English & OLID & 14,100 & T & offensive, non-offensive \\
{\em EN-SOLID} & English & SOLID & 120,758 & T & offensive, non-offensive \\
{\em HI} & Hindi & HASOC & 8,000 & T & hate offensive, non hate-offensive \\

\hline
\end{tabular}
}
\caption{Instances, sources, and labels in all datasets. F stands for Facebook and T for Twitter.}
\label{tab:data}
\end{table*}

\vspace{-4mm}

\paragraph{Other Datasets} In addition to the Marathi dataset, we used the four aforementioned publicly available offensive language detection datasets presented in Table \ref{tab:data}. OLID ({\em EN-OLID}) is one of the most popular offensive language datasets for English and we used its level A annotations (offensive vs. non-offensive) as labels.
We used {\em EN-SOLID}, the largest available dataset of its kind as our second English dataset. {\em EN-SOLID} contains over nine million English tweets labeled in a weakly supervised manner \cite{rosenthal2020}. {\em EN-SOLID} was created using an ensemble of four different models and provides, along with the class labels, the average and standard deviation of the confidence scores predicted by each model. We included only training examples with average confidence scores greater than 0.85 over all models, leaving us with 120,758 examples. Using both {\em EN-OLID} and {\em EN-SOLID} allows us to investigate the impact of training data size and help us answer {\bf RQ1}.

To perform transfer learning from a closely-related language to Marathi, we used {\em HI} \cite{mandl2019overview}.
Both the English and Hindi datasets contain Twitter data making them in-domain with respect to {\em MOLD}. {\em BE}, the Bengali dataset \cite{bhattacharya-etal-2020-developing}, 
is different than the other datasets as it contain Facebook data and three classes, allowing us to compare the performance of cross-lingual embeddings on off-domain data but in a a language similar to Marathi. For Bengali we merged the classes overtly aggressive and covertly aggressive and map them to {\em EN-OLID}'s offensive class. Using both {\em BE} and {\em HI} in addition to the two English datasets allow us to investigate the impact of language similarity aiming to answer our {\bf RQ2}.

\section{Methods and Results}
\label{sec:methodology}
\subsection{Monolingual Models}
We run several computational models on {\em MOLD}. We trained four classical machine learning classifiers, available in Scikit-learn \cite{scikit-learn}: Decision Trees, Naive Bayes, Random Forest, and SVM using bag of words (BoW), word unigrams, and word unigrams and bigrams combines using TF-IDF weighting. We took several pre-processing steps before extracting features such as removing numbers, extra spaces, special characters, and stop words.\footnote{Marathi stop words are available on \url{https://github.com/stopwords-iso/stopwords-mr/blob/master/stopwords-mr.txt}}


We implemented several deep learning models, such as multi layer perceptron (MLP), long short-term memory networks (LSTMs) with embedding layers, and bi-LSTMs with attention and word embedding layers. We used the Marathi word2vec embeddings released in \citet{kumar-etal-2020-passage}. We also experimented with several SOTA transformer models that support Marathi: multilingual BERT (BERT-m) \cite{devlin2019bert} and XLM-Roberta (XLM-R) \cite{conneau2019unsupervised}. XLM-R has an additional advantage: the embeddings are cross-lingual. This helps facilitate transfer learning across languages, as presented later in this section. We followed the same architecture described in \citet{ranasinghe-etal-2020-multilingual} where a simple softmax layer is added to the top of the classification (\textsc{[CLS]}) token to predict the probability of a class label. For XLM-R, from the available two pre-trained models, we specifically used the XLM-R large model.

\noindent For both classical and deep learning models we finetuned hyperparameters manually to obtain the best results for the validation set created using a 0.8:0.2 split on the training data. As the deep learning models tend to overfit, we evaluated the model on the validation set once in every 100 training batches. We performed \textit{early stopping} if the validation loss did not improve over 10 evaluation steps. All the deep learning experiments were run on an Nvidia Tesla K80 GPU.

\begin{figure}[ht]
\centering
\includegraphics[scale=0.45]{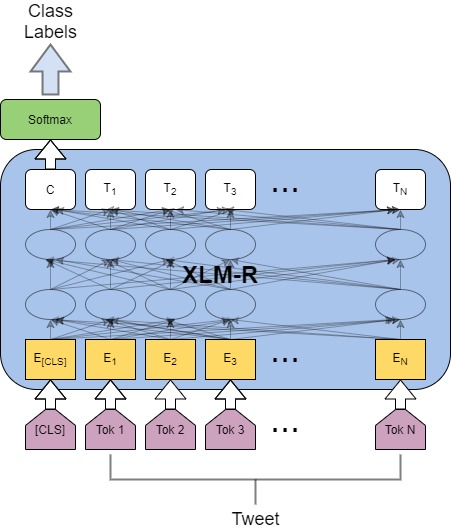}
\caption{Text classification architecture with XLM-R \cite{ranasinghe-etal-2020-multilingual}.}
\label{fig:architecture}
\end{figure}

\noindent Table \ref{tab:all} shows the results obtained by all monolingual models on {\em MOLD}'s test set in terms of both Macro F1 and Weighted F1. We use both metrics due to the data imbalance in {\em MOLD}. With the exception of MLP, all of the deep learning models outperformed the classical ones. This is somewhat surprising as classical models tend to outperform deep models on relatively small datasets like {\em MOLD}, but it corroborates the findings from recent competitions on this topic \cite{basile2019semeval}. 
Of the deep learning models, XLM-R transformers provided the best results with a 0.91 macro F1 score. 

\begin{table}[!ht]
\centering
\setlength{\tabcolsep}{4.5pt}
\scalebox{0.95}{
\begin{tabular}{llcc}
\hline
\bf Features & \bf Model   & \bf M F1 & \bf W F1 \\ \hline

Embeddings & XLM-R       & 0.9103   & 0.9210      \\
Embeddings & BERT-m      & 0.8852   & 0.8994       \\

Embeddings & LSTM & 0.8400   & 0.84.09        \\
Embeddings & Bi-LSTM & 0.8238   & 0.8251     \\

BoW & Random Forest & 0.7686   & 0.7796      \\
Embeddings & MLP & 0.7541   & 0.7830     \\
BoW & SVM & 0.7489   & 0.7813     \\

BoW & Naive Bayes & 0.7223   & 0.7597     \\
BoW & Decision Tree  & 0.7028   & 0.7395     \\

 \hline

\end{tabular}
}
\caption{Monolingual results for Marathi ordered by macro (M) F1. We also report weighted (W) F1 scores.}
\label{tab:all}
\end{table}

\subsection{Cross-lingual Models} 
The main appeal of transfer learning is its potential to leverage models trained on data from outside the domain of interest. This can be particularly helpful for boosting the performance of learning on low-resource languages like Marathi. The recent success of XLM-R cross-lingual transformers with transfer learning in offensive language identification for low resource languages \cite{ranasinghe-etal-2020-multilingual} confirms that this is a feasible approach. In these experiments, however, the transfer learning's base language was English whereas here we use two languages related to Marathi: Bengali and Hindi, in order to evaluate the extent to which language similarity boosts transfer learning performance. 

\paragraph{Transfer Learning} We first trained the XLM-R model separately on the {\em BE},  {\em HI},  {\em EN-OLID} and {\em EN-SOLID} datasets. Then we saved the weights of the transformer model and the softmax layer and used these weights to initialize the weights of the transformer-based classification model for Marathi.  TL row in Table \ref{tab:transfer} shows the results obtained by the cross lingual models with XLM-R. The use of transfer learning substantially improved the monolingual results. With 8,000 and 4,000 training instances, respectively, the transfer learning model achieved macro F1 scores of 0.9401 from Hindi  and of 0.9345 from Bengali, respectively, outperforming the results obtained using the two English datasets, {\em EN-OLID} and, especially, {\em EN-SOLID}, each contain more instances than either the Hindi or the Bengali dataset, yet they fail to outperform either as the base dataset in our transfer learning experiments, suggesting that language similarity played a positive role in transfer learning. 

\paragraph{Zero shot learning} To further observe the impact of language similarity in transfer learning, we performed Zero shot learning, where the XLM-R model was trained on the other datasets and tested on the Marathi test set. According to the results in Zero-shot row of Table \ref{tab:transfer} \textit{HI} outperforms all the other languages in Zero shot too. 

\begin{table}[!ht]
\centering
\setlength{\tabcolsep}{4.5pt}
\scalebox{0.95}{
\begin{tabular}{llcc}
\hline
\bf Methodology & \bf Dataset &   \bf M F1 & \bf W F1 \\ \hline

& HI &  0.9401   & 0.9492     \\
Transfer & BE &  0.9345   & 0.9422     \\
Learning & EN-SOLID & 0.9321   & 0.9399     \\
& EN-OLID &  0.9298   & 0.9385     \\
 \hline
 & HI &  0.8396   & 0.8461     \\
Zero-Shot & BE &  0.8115   & 0.8176     \\
& EN-SOLID & 0.7954   & 0.8004     \\
& EN-OLID &  0.7854   & 0.7901     \\
 \hline
\end{tabular}
}
\caption{Transfer learning results ordered by macro (M) F1 for Marathi. We also report weighted (W) F1 scores.}
\label{tab:transfer}
\end{table}

\begin{figure}[!ht]
\centering
\includegraphics[scale=0.55]{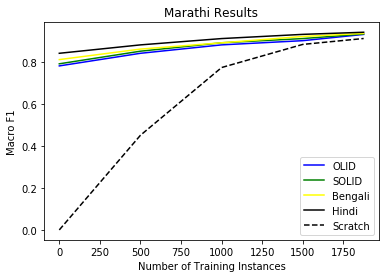}
\caption{Macro F1 with different number of examples and with different transfer learning strategies for Marathi}
\label{fig:progress}
\end{figure}

\paragraph{Few shot learning} Finally, we evaluated each of the languages performance in few shot learning with Marathi. We retrained offensive language identification XLM-R models from other languages on 100, 200, 300 etc. instances from Marathi. As shown in Figure \ref{fig:progress} \textit{HI} tops other languages in all the few shot experiments making it further clear that transfer learning from a more similar language is effective in offensive language identification.

\section{Conclusion and Future Work}

This paper introduced {\em MOLD}, the first offensive language dataset for Marathi. We evaluated the performance of several machine learning models trained to identify offensive content in Marathi. Our results show that applying cross-lingual contextual word embeddings substantially improved performance over monolingual models. Furthermore, we showed that XLM-R with transfer learning from Hindi outperforms all of the other methods we tested. The results obtained by our models confirm that closely related languages provide an advantage in our transfer learning experiments, answering our {\bf RQ2}. This is likely due to the fact that Hindi and Marathi are typologically related and also because these languages are in a situation of language contact sharing cultural background. 

To the best of our knowledge, this paper is the first to address the question of language similarity in cross-lingual learning for offensive language identification. With respect to our {\bf RQ1}, our results show that the difference in performance between transfer learning strategies from OLID and from SOLID is minimal. SOLID is more than eight times larger than OLID, suggesting that beyond a certain point, more instances do not necessarily yield significant performance improvements in transfer learning. Finally, we believe that the findings presented in this paper can open a wide range of avenues to offensive language identification applied to other low resource languages, particularly from the Indo-Aryan family. 

{\em MOLD} is the official dataset for Marathi at the HASOC 2021\footnote{\url{https://hasocfire.github.io/hasoc/2021/index.html}} shared task on Hate Speech and Offensive Content Identification in English and Indo-Aryan Languages. We are expanding the annotation of this dataset to the levels B and C of OLID's annotation taxonomy. This will provide us with the opportunity to test computational models to identify the type and target of offensive posts in Marathi. As future work, we would like to evaluate the performance of transfer learning from Dravidian languages spoken in India such as Tamil and Telugu to analyze the interplay between language similarity and cultural overlap in cross-lingual offensive language identification as in \newcite{ranasinghe2021evaluation}. 




\section*{Acknowledgment}

We would like to thank the dataset annotators who helped us with the annotation of {\em MOLD}. We further thank the anonymous RANLP reviewers for the insightful feedback provided. 

\bibliographystyle{acl_natbib}
\bibliography{ranlp2021}

\begin{thebibliography}{42}
\expandafter\ifx\csname natexlab\endcsname\relax\def\natexlab#1{#1}\fi

\bibitem[{Plaza-del Arco et~al.(2021)Plaza-del Arco, Molina-Gonz{\'a}lez,
  Ure{\~n}a-L{\'o}pez, and Mart{\'\i}n-Valdivia}]{plaza2021comparing}
Flor~Miriam Plaza-del Arco, M~Dolores Molina-Gonz{\'a}lez, L~Alfonso
  Ure{\~n}a-L{\'o}pez, and M~Teresa Mart{\'\i}n-Valdivia. 2021.
\newblock Comparing pre-trained language models for spanish hate speech
  detection.
\newblock \emph{Expert Systems with Applications}, 166:114120.

\bibitem[{Aroyehun and Gelbukh(2018)}]{aroyehun2018aggression}
Segun~Taofeek Aroyehun and Alexander Gelbukh. 2018.
\newblock Aggression detection in social media: Using deep neural networks,
  data augmentation, and pseudo labeling.
\newblock In \emph{Proceedings of TRAC}.

\bibitem[{Basile et~al.(2019)Basile, Bosco, Fersini, Nozza, Patti, Pardo,
  Rosso, and Sanguinetti}]{basile2019semeval}
Valerio Basile, Cristina Bosco, Elisabetta Fersini, Debora Nozza, Viviana
  Patti, Francisco Manuel~Rangel Pardo, Paolo Rosso, and Manuela Sanguinetti.
  2019.
\newblock Semeval-2019 task 5: Multilingual detection of hate speech against
  immigrants and women in twitter.
\newblock In \emph{Proceedings of SemEval}.

\bibitem[{Bhattacharya et~al.(2020)Bhattacharya, Singh, Kumar, Bansal, Bhagat,
  Dawer, Lahiri, and Ojha}]{bhattacharya-etal-2020-developing}
Shiladitya Bhattacharya, Siddharth Singh, Ritesh Kumar, Akanksha Bansal, Akash
  Bhagat, Yogesh Dawer, Bornini Lahiri, and Atul~Kr. Ojha. 2020.
\newblock Developing a multilingual annotated corpus of misogyny and
  aggression.
\newblock In \emph{Proceedings of TRAC}.

\bibitem[{Bohra et~al.(2018)Bohra, Vijay, Singh, Akhtar, and
  Shrivastava}]{bohra2018dataset}
Aditya Bohra, Deepanshu Vijay, Vinay Singh, Syed~Sarfaraz Akhtar, and Manish
  Shrivastava. 2018.
\newblock A dataset of hindi-english code-mixed social media text for hate
  speech detection.
\newblock In \emph{Proceedings of PEOPLES}.

\bibitem[{Bucur et~al.(2021)Bucur, Zampieri, and
  Dinu}]{bucur-etal-2021-exploratory}
Ana-Maria Bucur, Marcos Zampieri, and Liviu~P. Dinu. 2021.
\newblock An exploratory analysis of the relation between offensive language
  and mental health.
\newblock In \emph{Findings of the ACL}.

\bibitem[{Carletta(1996)}]{carletta1996assessing}
Jean Carletta. 1996.
\newblock Assessing agreement on classification tasks: The kappa statistic.
\newblock \emph{Computational Linguistics}, 22(2):249--254.

\bibitem[{Caselli et~al.(2020)Caselli, Basile, Mitrovi{\'c}, and
  Granitzer}]{caselli2020hatebert}
Tommaso Caselli, Valerio Basile, Jelena Mitrovi{\'c}, and Michael Granitzer.
  2020.
\newblock Hatebert: Retraining bert for abusive language detection in english.
\newblock \emph{arXiv preprint arXiv:2010.12472}.

\bibitem[{\c{C}\"{o}ltekin(2020)}]{coltekin2020}
\c{C}a\u{g}r{\i} \c{C}\"{o}ltekin. 2020.
\newblock {A Corpus of Turkish Offensive Language on Social Media}.
\newblock In \emph{Proceedings of LREC}.

\bibitem[{Chiril et~al.(2019)Chiril, Benamara~Zitoune, Moriceau, Coulomb-Gully,
  and Kumar}]{chiril-etal-2019-multilingual}
Patricia Chiril, Farah Benamara~Zitoune, V{\'e}ronique Moriceau, Marl{\`e}ne
  Coulomb-Gully, and Abhishek Kumar. 2019.
\newblock Multilingual and multitarget hate speech detection in tweets.
\newblock In \emph{Proceedings of TALN}.

\bibitem[{Conneau et~al.(2019)Conneau, Khandelwal, Goyal, Chaudhary, Wenzek,
  Guzm{\'a}n, Grave, Ott, Zettlemoyer, and Stoyanov}]{conneau2019unsupervised}
Alexis Conneau, Kartikay Khandelwal, Naman Goyal, Vishrav Chaudhary, Guillaume
  Wenzek, Francisco Guzm{\'a}n, Edouard Grave, Myle Ott, Luke Zettlemoyer, and
  Veselin Stoyanov. 2019.
\newblock Unsupervised cross-lingual representation learning at scale.
\newblock In \emph{Proceedings of ACL}.

\bibitem[{Dadvar et~al.(2013)Dadvar, Trieschnigg, Ordelman, and
  de~Jong}]{dadvar2013improving}
Maral Dadvar, Dolf Trieschnigg, Roeland Ordelman, and Franciska de~Jong. 2013.
\newblock {Improving Dyberbullying Detection with User Context}.
\newblock In \emph{Proceedings of ECIR}.

\bibitem[{Devlin et~al.(2019)Devlin, Chang, Lee, and
  Toutanova}]{devlin2019bert}
Jacob Devlin, Ming-Wei Chang, Kenton Lee, and Kristina Toutanova. 2019.
\newblock {BERT: Pre-training of Deep Bidirectional Transformers for Language
  Understanding}.
\newblock In \emph{Proceedings of NAACL}.

\bibitem[{Fi\v{s}er et~al.(2017)Fi\v{s}er, Erjavec, and
  Ljube\v{s}i\'{c}}]{fiser2017}
Darja Fi\v{s}er, Toma\v{z} Erjavec, and Nikola Ljube\v{s}i\'{c}. 2017.
\newblock {Legal Framework, Dataset and Annotation Schema for Socially
  Unacceptable On-line Discourse Practices in Slovene}.
\newblock In \emph{Proceedings ALW}.

\bibitem[{Fortuna et~al.(2019)Fortuna, da~Silva, Wanner, Nunes
  et~al.}]{fortuna2019hierarchically}
Paula Fortuna, Joao~Rocha da~Silva, Leo Wanner, S{\'e}rgio Nunes, et~al. 2019.
\newblock {A Hierarchically-labeled Portuguese Hate Speech Dataset}.
\newblock In \emph{Proceedings of ALW}.

\bibitem[{Ghadery and Moens(2020)}]{ghadery-moens-2020-liir}
Erfan Ghadery and Marie-Francine Moens. 2020.
\newblock {LIIR} at {S}em{E}val-2020 task 12: A cross-lingual augmentation
  approach for multilingual offensive language identification.
\newblock In \emph{Proceedings of SemEval}.

\bibitem[{Hettiarachchi and
  Ranasinghe(2019)}]{hettiarachchi-ranasinghe-2019-emoji}
Hansi Hettiarachchi and Tharindu Ranasinghe. 2019.
\newblock Emoji powered capsule network to detect type and target of offensive
  posts in social media.
\newblock In \emph{Proceedings of RANLP}.

\bibitem[{Kumar et~al.(2018)Kumar, Ojha, Malmasi, and
  Zampieri}]{kumar2018benchmarking}
Ritesh Kumar, Atul~Kr Ojha, Shervin Malmasi, and Marcos Zampieri. 2018.
\newblock {Benchmarking Aggression Identification in Social Media}.
\newblock In \emph{Proceedings of TRAC}.

\bibitem[{Kumar et~al.(2020{\natexlab{a}})Kumar, Ojha, Malmasi, and
  Zampieri}]{kumar-etal-2020-evaluating}
Ritesh Kumar, Atul~Kr. Ojha, Shervin Malmasi, and Marcos Zampieri.
  2020{\natexlab{a}}.
\newblock Evaluating aggression identification in social media.
\newblock In \emph{Proceedings of TRAC}.

\bibitem[{Kumar et~al.(2020{\natexlab{b}})Kumar, Kumar, Kanojia, and
  Bhattacharyya}]{kumar-etal-2020-passage}
Saurav Kumar, Saunack Kumar, Diptesh Kanojia, and Pushpak Bhattacharyya.
  2020{\natexlab{b}}.
\newblock {``}a passage to {I}ndia{''}: Pre-trained word embeddings for
  {I}ndian languages.
\newblock In \emph{Proceedings of SLTU}.

\bibitem[{Liu et~al.(2019)Liu, Li, and Zou}]{liu-etal-2019-nuli}
Ping Liu, Wen Li, and Liang Zou. 2019.
\newblock {NULI} at {S}em{E}val-2019 task 6: Transfer learning for offensive
  language detection using bidirectional transformers.
\newblock In \emph{Proceedings of SemEval}.

\bibitem[{Malmasi and Zampieri(2017)}]{malmasi2017detecting}
Shervin Malmasi and Marcos Zampieri. 2017.
\newblock {Detecting Hate Speech in Social Media}.
\newblock In \emph{Proceedings of RANLP}.

\bibitem[{Mandl et~al.(2019)Mandl, Modha, Majumder, Patel, Dave, Mandlia, and
  Patel}]{mandl2019overview}
Thomas Mandl, Sandip Modha, Prasenjit Majumder, Daksh Patel, Mohana Dave,
  Chintak Mandlia, and Aditya Patel. 2019.
\newblock Overview of the hasoc track at fire 2019: Hate speech and offensive
  content identification in indo-european languages.
\newblock In \emph{Proceedings of FIRE}.

\bibitem[{Mubarak et~al.(2021)Mubarak, Rashed, Darwish, Samih, and
  Abdelali}]{mubarak2020arabic}
Hamdy Mubarak, Ammar Rashed, Kareem Darwish, Younes Samih, and Ahmed Abdelali.
  2021.
\newblock {Arabic Offensive Language on Twitter: Analysis and Experiments}.
\newblock In \emph{Proceedings of WANLP}.

\bibitem[{Pamungkas and Patti(2019)}]{pamungkas2019cross}
Endang~Wahyu Pamungkas and Viviana Patti. 2019.
\newblock Cross-domain and cross-lingual abusive language detection: A hybrid
  approach with deep learning and a multilingual lexicon.
\newblock In \emph{Proceedings ACL:SRW}.

\bibitem[{Pedregosa et~al.(2011)Pedregosa, Varoquaux, Gramfort, Michel,
  Thirion, Grisel, Blondel, Prettenhofer, Weiss, Dubourg, Vanderplas, Passos,
  Cournapeau, Brucher, Perrot, and Duchesnay}]{scikit-learn}
F.~Pedregosa, G.~Varoquaux, A.~Gramfort, V.~Michel, B.~Thirion, O.~Grisel,
  M.~Blondel, P.~Prettenhofer, R.~Weiss, V.~Dubourg, J.~Vanderplas, A.~Passos,
  D.~Cournapeau, M.~Brucher, M.~Perrot, and E.~Duchesnay. 2011.
\newblock Scikit-learn: Machine learning in {P}ython.
\newblock \emph{Journal of Machine Learning Research}, 12.

\bibitem[{Pitenis et~al.(2020)Pitenis, Zampieri, and Ranasinghe}]{pitenis2020}
Zeses Pitenis, Marcos Zampieri, and Tharindu Ranasinghe. 2020.
\newblock {Offensive Language Identification in Greek}.
\newblock In \emph{Proceedings of LREC}.

\bibitem[{Poletto et~al.(2017)Poletto, Stranisci, Sanguinetti, Patti, and
  Bosco}]{poletto2017hate}
Fabio Poletto, Marco Stranisci, Manuela Sanguinetti, Viviana Patti, and
  Cristina Bosco. 2017.
\newblock {Hate Speech Annotation: Analysis of an Italian Twitter Corpus}.
\newblock In \emph{Proceedings of CLiC-it}.

\bibitem[{Ranasinghe and Zampieri(2020)}]{ranasinghe-etal-2020-multilingual}
Tharindu Ranasinghe and Marcos Zampieri. 2020.
\newblock {{Multilingual Offensive Language Identification with Cross-lingual
  Embeddings}}.
\newblock In \emph{Proceedings of EMNLP}.

\bibitem[{Ranasinghe and
  Zampieri(2021{\natexlab{a}})}]{ranasinghe2021evaluation}
Tharindu Ranasinghe and Marcos Zampieri. 2021{\natexlab{a}}.
\newblock {An Evaluation of Multilingual Offensive Language Identification
  Methods for the Languages of India}.
\newblock \emph{Information}, 12(8):306.

\bibitem[{Ranasinghe and Zampieri(2021{\natexlab{b}})}]{ranasinghemudes}
Tharindu Ranasinghe and Marcos Zampieri. 2021{\natexlab{b}}.
\newblock {MUDES: Multilingual Detection of Offensive Spans}.
\newblock In \emph{Proceedings of NAACL}.

\bibitem[{Ranasinghe and Zampieri(2021{\natexlab{c}})}]{ranasingheTALLIP}
Tharindu Ranasinghe and Marcos Zampieri. 2021{\natexlab{c}}.
\newblock {Multilingual Offensive Language Identification for Low-resource
  Languages}.
\newblock \emph{ACM Transactions on Asian and Low-Resource Language Information
  Processing (TALLIP)}.

\bibitem[{Ranasinghe et~al.(2019)Ranasinghe, Zampieri, and
  Hettiarachchi}]{ranasinghe2019brums}
Tharindu Ranasinghe, Marcos Zampieri, and Hansi Hettiarachchi. 2019.
\newblock {BRUMS at HASOC 2019: Deep Learning Models for Multilingual Hate
  Speech and Offensive Language Identification}.
\newblock In \emph{Proceedings of FIRE}.

\bibitem[{Ridenhour et~al.(2020)Ridenhour, Bagavathi, Raisi, and
  Krishnan}]{ridenhour2020detecting}
Michael Ridenhour, Arunkumar Bagavathi, Elaheh Raisi, and Siddharth Krishnan.
  2020.
\newblock Detecting online hate speech: Approaches using weak supervision
  and network embedding models.
\newblock In \emph{Proceedings of SBP-BRiMS}.

\bibitem[{Rosenthal et~al.(2021)Rosenthal, Atanasova, Karadzhov, Zampieri, and
  Nakov}]{rosenthal2020}
Sara Rosenthal, Pepa Atanasova, Georgi Karadzhov, Marcos Zampieri, and Preslav
  Nakov. 2021.
\newblock {SOLID: A Large-Scale Weakly Supervised Dataset for Offensive
  Language Identification}.
\newblock In \emph{Findings of the ACL}.

\bibitem[{Sai and Sharma(2021)}]{sai-sharma-2021-towards}
Siva Sai and Yashvardhan Sharma. 2021.
\newblock Towards offensive language identification for {D}ravidian languages.
\newblock In \emph{Proceedings of DravidianLangTech}.

\bibitem[{Tulkens et~al.(2016)Tulkens, Hilte, Lodewyckx, Verhoeven, and
  Daelemans}]{tulkens2016dictionary}
St{\'e}phan Tulkens, Lisa Hilte, Elise Lodewyckx, Ben Verhoeven, and Walter
  Daelemans. 2016.
\newblock {A Dictionary-based Approach to Racism Detection in Dutch Social
  Media}.
\newblock In \emph{Proceedings of TA-COS}.

\bibitem[{Wiegand et~al.(2018)Wiegand, Siegel, and
  Ruppenhofer}]{wiegand2018overview}
Michael Wiegand, Melanie Siegel, and Josef Ruppenhofer. 2018.
\newblock Overview of the {GermEval} 2018 shared task on the identification of
  offensive language.
\newblock In \emph{Proceedings of GermEval}.

\bibitem[{Yang et~al.(2019)Yang, Dai, Yang, Carbonell, Salakhutdinov, and
  Le}]{NEURIPS2019_dc6a7e65}
Zhilin Yang, Zihang Dai, Yiming Yang, Jaime Carbonell, Russ~R Salakhutdinov,
  and Quoc~V Le. 2019.
\newblock {XLNet: Generalized Autoregressive Pretraining for Language
  Understanding}.
\newblock In \emph{Proceedings of NeurIPS}.

\bibitem[{Yao et~al.(2019)Yao, Chelmis, and Zois}]{yao2019cyberbullying}
Mengfan Yao, Charalampos Chelmis, and Daphney-Stavroula Zois. 2019.
\newblock {Cyberbullying Ends Here: Towards Robust Detection of Cyberbullying
  in Social Media}.
\newblock In \emph{Proceedings of WWW}.

\bibitem[{Zampieri et~al.(2019)Zampieri, Malmasi, Nakov, Rosenthal, Farra, and
  Kumar}]{OLID}
Marcos Zampieri, Shervin Malmasi, Preslav Nakov, Sara Rosenthal, Noura Farra,
  and Ritesh Kumar. 2019.
\newblock Predicting the type and target of offensive posts in social media.
\newblock In \emph{Proceedings of NAACL}.

\bibitem[{Zampieri et~al.(2020)Zampieri, Nakov, Rosenthal, Atanasova,
  Karadzhov, Mubarak, Derczynski, Pitenis, and
  \c{C}\"{o}ltekin}]{zampieri-etal-2020-semeval}
Marcos Zampieri, Preslav Nakov, Sara Rosenthal, Pepa Atanasova, Georgi
  Karadzhov, Hamdy Mubarak, Leon Derczynski, Zeses Pitenis, and
  \c{C}a\u{g}r{\i} \c{C}\"{o}ltekin. 2020.
\newblock {SemEval-2020 Task 12: Multilingual Offensive Language Identification
  in Social Media (OffensEval 2020)}.
\newblock In \emph{Proceedings of SemEval}.

\end{thebibliography}

\end{document}